\documentclass[conference]{IEEEtran}
\usepackage{cite}
\usepackage{graphicx}
\graphicspath{ {images/} }

\begin{document}


\title{Quality Assessment Metrics for Edge Detection and Edge-aware Filtering: A Tutorial Review}


\author
{\IEEEauthorblockN{Diana Sadykova, and Alex Pappachen James}
\IEEEauthorblockA{School of Engineering\\
Nazarbayev University, Astana\\
Email: Diana.Sadykova@nu.edu.kz; apj@ieee.org }
}
\maketitle

\begin{abstract}

The quality assessment of edges in an image is an important topic as it helps to benchmark the performance of edge detectors, and edge-aware filters that are used in a wide range of image processing tasks. The most popular image quality metrics such as Mean squared error (MSE), Peak signal-to-noise ratio (PSNR) and Structural similarity (SSIM) metrics for assessing and justifying the quality of edges. However, they do not address the structural and functional accuracy of edges in images with a wide range of natural variabilities. In this review, we provide an overview of all the most relevant performance metrics that can be used to benchmark the quality performance of edges in images. We identify four major groups of metrics and also provide a critical insight into the evaluation protocol and governing equations.

\end{abstract}

\begin {IEEEkeywords}
Edge preservation, image metrics, edge quality measures, edge detection, sobel filters, mean square error, PSNR, SSIM
\end{IEEEkeywords}

\section{Introduction}
Edges in the images from the most basic visual cues for the identification and recognition objects.
Application of conventional image filters results in blurring of the edges due to inhomogeneity in pixel intensity values. Various edge aware filters have been introduced as a promising solution, including anisotropic diffusion, bilateral filter, domain transform filter, guided filter, wavelet transform based filtering and G-neighbor classification filtering \cite{survey, ad, bm, gm,gn}. It is commonly accepted that efficiency of the filters is graded according to Mean squared error (MSE), Peak signal-to-noise ratio (PSNR) and Structural similarity index (SSIM) metrics \cite{ad2}. However, these types of assessment orients only on overall image quality and therefore are not completely suitable for understanding the actual quality of edges. In this work, we provide a critical review of most promising edge-oriented metrics suitable for analysis and measurement of edge quality.

There are two main groups of image quality assessment for edges: subjective and objective. Subjective evaluation can be represented by mean opinion score (MOS) or an average of human visual assessment. This type of the evaluation is considered to be more exact and accurate\cite{zhang14}. The second approach is applied compared with a set standards and utilization of an original or an ideal image. Objective techniques are divided into three categories: (1) full reference (FR), (2) reduced reference (RR) and (3) no reference (NR) evaluation. The FR evaluation results in a better performance, as the whole source of information is provided, however, this might lead to vast data transmission. RR comparison can solve this problem, but regions of interests should be carefully selected\cite{carnec5}. The least used method is NR because it is entirely based on information from a distorted image \cite{lan15}. 

Commonly used metrics are MSE, PSNR and SSIM. MSE is basically a weighted function of deviations in images, or square difference between compared images \cite{huynh16}. In Eq. \ref{Eq::MSE}, $M$ and $N$ stands for image size, while $I_1(s,t)$ and $I_2(s,t)$ for locations.

\begin{equation}
MSE=\frac{1}{MN} \sum_{i=1}^{M} \sum_{j=1}^{N} (I_1(i,j)-I_2(i,j))^2
\label{Eq::MSE}
\end{equation}

Another measure, which is strongly related to MSE is PSNR, defined by Eq. \ref{Eq::PSNR}. It indicates the level of losses or signals integrity \cite{wang17}.

\begin{equation}
PSNR = 10log (\frac{max(I)^2}{MSE}) 
\label{Eq::PSNR}
\end{equation}

Although these measures are widely used, experimental results showed, that images with a different level of distortion might result in similar PSNR and MSE values \cite{gcmse}.

SSIM is more closely related to the human visual system as it extracts useful information as luminance (l), contrast (c) and structure (s). It can be applied to evaluate structure preservation and noise removal \cite{md18}.

\begin{equation}
SSIM= function (l (I_1,I_2),c (I_1,I_2),s (I_1,I_2)) 
\label{Eq::SSIM}
\end{equation}

The main limitation of SSIM measure is inability to measure highly blurred images \cite{essim} successfully. All the three most common metrics MSE, PSNR, and SSIM, are limited in their use for benchmarking the performance of edge in the images.

\section{Edge-aware performance metrics}

Figure 1 shows the overview of the metrics that can be useful for benchmarking the performance of edge quality of objects in the images after edge detection or after applying an edge-aware filter. Overall, we can group the metrics into four distinct categories, i.e. that based on (a) Sobel filtering, (2) SSIM, (3) MSE and (4) Gaussian kernel functions.

\begin{figure}[h]
\centering
\includegraphics[width=90mm]{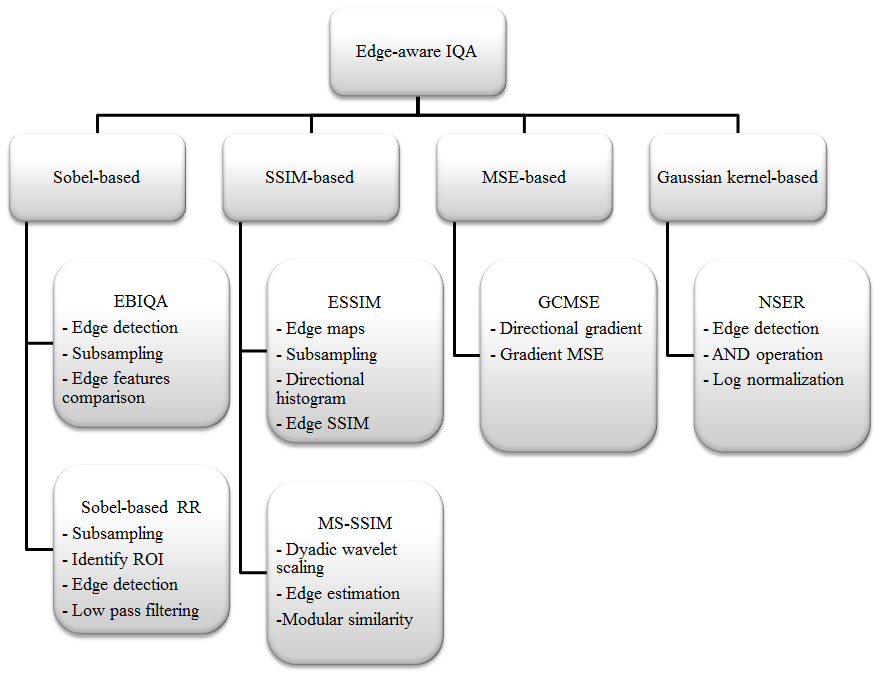}\\
\caption{Overview of edge based IQA}
\label{fig::eiqa}
\end{figure}

\subsection{EBIQA: Edge based image quality assessment}

Edge preservation one of the most important aspect during the human visual assessment. Attar, Shahbahrami, and Rad proposed EBIQA technique that aims to operate on the human perception of the features \cite{ebiqa}. The metric consists of the four major steps, where original and distorted images are compared:
\begin{enumerate}
    \item Edges locations are identified utilizing Sobel edge detector technique in both images.
    \item The 16 $\times$ 16 pixel window size vectors are formed at each image based on Eq.\ref{Eq::EBIQA} and Eq. \ref{Eq::EBIQA2}, where $I_1$ is reference image and $I_2$ is the image under test \cite{ebiqa2}.
\begin{equation}
I_1=(O,AL,PL,N,VHO)
\label{Eq::EBIQA}
\end{equation}
\begin{equation}
I_2=(O,AL,PL,N,VHO)
\label{Eq::EBIQA2}
\end{equation}

Here, each vector uses information about five edge oriented characteristics:
\begin{itemize}
\item 'O' stands for edge orientation in the image, which is a total number of edges.
\item 'AL' or value of the average length of all edges. 
\item 'PL' estimates the number of pixels with a similar level of intensity values. 
\item 'N' responds for sum of pixels, which form edges. 
\item 'VH' corresponds for sum of pixels, which form edges in either vertically or horizontally located edges.
\end {itemize}

Note that initially in 2011 authors proposed only four edge related features for vector creation, but later in 2016 one more criterion was included for better performance, which is 'VH' \cite{ebiqa2}.

    \item Finally, we estimate EBIQA by Eq. \ref{Eq::EBIQA4}, where average Euclidean distance of proposed vectors is estimated.
    
 \begin{equation}
EBIQA=\frac{1}{MN} \sum_{i=1}^{M} \sum_{j=1}^{N}\sqrt{(I_1-I_2)^2}
\label{Eq::EBIQA4}
\end{equation}

\end{enumerate}

\paragraph*{Merits} 
The main merits of using the EBIQA metric are:

 \begin{itemize}
\item Application of Sobel filter considered to be time efficient, effective and simple.  
\item Provides better performance than PSNR and SSIM in FR evaluation.
\item Assessment of edge similarity showed results similar to MOS i.e. had a good response to image degradation.
\item Might be further utilized for RR or NR.
\end {itemize}

\paragraph*{Demerits} 
\begin{itemize}
\item The need to have large data transmission .
\end {itemize}

\subsection{Sobel-based reduced reference evaluation}

The work which was done by Martini et. al \cite{sobel} also draws upon the estimation of change in edge-based features. Similar to EBIQA Sobel filtering is applied  for edge identification \cite{ebiqa}. In this filter, weighted pixel difference equation is applied for 3$\times$3 windows, and then the pixel is considered to be an edge if it is bigger than a given threshold. The overall algorithm for the implementation includes: 
\begin{enumerate}
    \item The image is sub-sampled to have 16$\times$16 blocks.
    \item Then a selection of 12 out of all blocks is done using region of interest with centered symmetry as it is illustrated in Fig.\ref{fig::sobel}.
    
\begin{figure}[h]
\centering
\includegraphics[width=60mm]{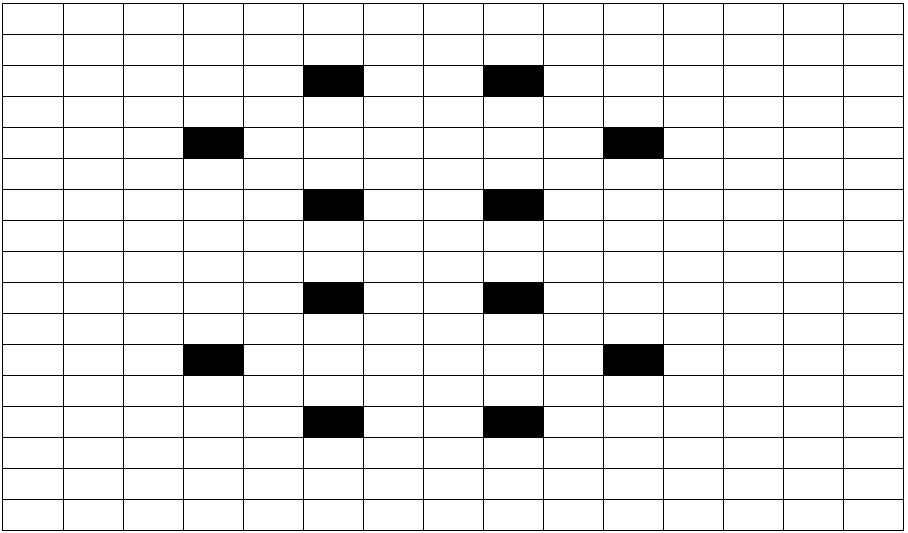}\\
\caption{Example of 12 blocks considered for RR evaluation}
\label{fig::sobel}
\end{figure}
    \item Sobel filter is applied separately to each of the selected block, rather than to the full image as in the case of EBIQA.
    \item Number of edges evaluated by low pass filtering edges by application of thresholds, referred as $t_1$ and $t_2$. Then the proposed parameter is estimated using Eq. \ref{Eq::sobel}, where $w_1$,$w_2$ are constant and $M\times N$ represent the total number of pixels.
 \begin{equation}
I=\frac{1}{MN} \sum_{i=1}^{MN} \frac{w_1 t_1+w_2 t_{1,2}}{MN}
\label{Eq::sobel}
\end{equation}

\end{enumerate}

\paragraph*{Merits} 
The main merits of using the Sobel based metric, they are: 

\begin{itemize}
\item Application of Sobel filter is considered to be time efficient, effective and simple.  
\item Provides better performance using RR, than PSNR and MSSIM in FR evaluation.
\item The implementation of equations requires only simplistic estimations.
\item This metric can be used for real-time estimations.
\end {itemize}

\paragraph*{Demerits} 
There are however some limitations of using the Sobel based metric, they are:
\begin{itemize}
\item Block size for each image should be separately regulated.
\item Threshold selection can be inappropriate and requires careful processing.
\item Transmission operation is time-consuming.
\end {itemize}

\subsection{ESSIM: Edge based structural similarity}
The IQA technique proposed by Chen et al. is the edge oriented version of SSIM metric. Mainly, the structural similarity components in Eq. \ref{Eq::SSIM} is substituted by edge similarity component, see Eq. \ref{Eq::ESSIM} \cite{essim}. 
\begin{equation}
ESSIM= function (l (I_1,I_2),c (I_1,I_2),e (I_1,I_2)) 
\label{Eq::ESSIM}
\end{equation}

The following steps do the computation of this metric:
\begin{enumerate}
\item Vertical and Horizontal maps are created by the use of Sobel operator.
\item Image is sub-sampled to 16$\times$16 blocks.
\item Histogram of edge direction is created according to the sum of amplitudes with similar (1/8) directions.
\item Using standard deviations of obtained histograms the edge factor is calculated in Eq. \ref{Eq::ESSIM2}.

\begin{equation}
e (I_1,I_2)=\frac{\sigma_{I1 I2}+c_3}{\sigma_{I1}\sigma_{I2}+c_3}
\label{Eq::ESSIM2}
\end{equation}
\end{enumerate}

\paragraph*{Merits} 
The main advantages of using the ESSIM metric are: 
\begin{itemize}
\item Application of Sobel filter is considered to be time efficient, effective and simple.  
\item Provides better performance than PSNR and SSIM in FR evaluation.
\item Assessment of edge similarity showed results similar to MOS i.e. had a good response to image degradation.
\end {itemize}

\paragraph*{Demerits} 
The limitations of ESSIM metric are: 
\begin{itemize}
\item Prediction sequence sometimes shows poor output.
\item Efficiency depends on noise type applied.
\item Requires careful optimization of parameters.
\end {itemize}

Note, that in 2008 Cui and Allen in their work presented three SSIM measures, based on corner, edge, and symmetry. The ESSIM demonstrated the best performance among all of them \cite{essim2}. Also, it can be noticed, that first two steps of this metric are similar to the EBIQA, only the mathematical approaches are different when it comes to implementation.

\subsection{MS-SSIM: Multi scale structural similarity}
Another approach suggested by Zhai et.al also uses SSIM metric as a foundation \cite{msssim}. The difference is that the dyadic wavelet transform is used instead of Sobel filtering. The main steps of the metric calculation are:

\begin{enumerate}
\item The image is scaled using dyadic wavelet transform in vertical and horizontal directions. 
\item Then the edge 'distance' and 'angle' are estimated similar to the ESSIM.
\item Modular similarity and maximum modular similarity are estimated.
\end{enumerate}

\paragraph*{Merits} 
\begin{itemize}
\item Application of wavelet transform is shown to produce accurate results.
\item Provides better performance than PSNR and SSIM in FR and RR evaluation.
\end {itemize}

\paragraph*{Demerits} 
\begin{itemize}
\item The need to have large transmission data is the main limitation of using this metric.
\end {itemize}

Note, maximum modular similarity requires a higher level of scaling.

\subsection{NSER: Non-shift edge based ratio}

This method is based on zero-crossings and was proposed by Zhang, Mou and Zhang \cite{nser,nser2}. It makes decisions upon edge maps. It can be noted that this method combines some of the steps of previously mentioned metrics. The main stages for calculation of the metric are:
\begin{enumerate}
\item Gaussian kernel is applied to the interesting images on different standard deviation scales to identify edges.
\item Then operation is performed between two images. Ratio of the common edge number located by initial edge number is found using Eq. \ref{Eq::NSER}:
\begin{equation}
p_i=||I_1 \cap I_2||/||I_1||
\label{Eq::NSER}
\end{equation}

\item The result of this operation is normalized by log function to improve correlation factor:

\begin{equation}
NSER(I_1,I_2)=-\sum_{i=1}^{N}log_{10}(1-p_i)
\label{Eq::NSER2}
\end{equation}

\end{enumerate}

\paragraph*{Merits} 
\begin{itemize}
\item Provides similar performance as MS-SSIM in FR evaluation.
\item The metric is simplistic and easy to implement for real-time applications.
\end {itemize}

\paragraph*{Demerits} 
\begin{itemize}
\item This metric is not suitable for images having high levels distortion.
\item Efficiency varies from one database to another. 
\item Suitable normalization will be required to ensure robustness.
\end {itemize}

\subsection{GCMSE: Gradient conduction mean square error}

Lopez-Randulfe et al. in their paper introduced an edge-aware metric based on MSE \cite{gcmse}. In this algorithm weighted sum of gradients (distance pixels) is taken into account. The main steps for calculating the metric are:

\begin{enumerate}
\item Directional gradients are estimated in four directions using Eq. \ref{Eq::GCMSE2} and then average value $G_p$ is found. Note that the results are optimized by the coefficient $k$:

\begin{equation}
G=\frac{(I_2-I_1)^2}{(I_2-I_1)^2+k^2}
\label{Eq::GCMSE2}
\end{equation}

\item The GCMSE is estimated based on Eq. \ref{Eq::GCMSE}:

\begin{equation}
GCMSE=\frac{\sum_{x=1}^{m} \sum_{y=1}^{n} [(I_2(x,y)-I_1(x,y))G_p]^2}{C1+\sum_{x=1}^{m} \sum_{y=1}^{n} G_p} 
\label{Eq::GCMSE}
\end{equation}

\end{enumerate}

\paragraph*{Merits} 
\begin{itemize}
\item Provides better performance than MSE and SSIM in FR evaluation
\item The measure is straightforward and easy to implement.
\end {itemize}

\paragraph*{Demerits} 
\begin{itemize}
\item The parameters require careful optimization.
\item Multiple tests on a different set of databases are required to ensure robustness of results.
\item Needs normalization according to the size of the image used.
\end {itemize}


\section{Conclusion}

In this review paper, we presented the set of image quality metrics that are useful for the assessment of the quality of edges in images. These measures can be used to detect the robustness of edge detectors and edge-aware filters under noisy conditions. We do note that a single measure to assess the quality of edges is not sufficient to ensure the accuracy of the quality assessments and interpretations. Every metric has its limitations and challenges, making it necessary to use a multitude of measures specific to a given application and condition. Further, it has been observed that substantially all six mentioned metrics tend to provide results closer to human visual assessment than traditionally obtained ones  using PSNR and SSIM.

\bibliographystyle{unsrt}
\bibliography{bibliography}

\end{document}